\documentclass[11pt,letterpaper]{article}
\usepackage{emnlp2016}
\usepackage{times}
\usepackage{graphicx}
\usepackage{subcaption}
\usepackage{latexsym}
\usepackage{amsfonts}
\usepackage{amsmath}
\usepackage{multirow}
\usepackage[hyphens]{url}
\usepackage{comment}
\usepackage[inline]{enumitem}
\usepackage{color}
\usepackage[dvipsnames]{xcolor}

\newcommand\todo[1]{\textcolor{red}{\textbf{#1}}}
\emnlpfinalcopy 

\title{Conversational Contextual Cues:\\ The Case of Personalization and History for Response Ranking}

\author{Rami Al-Rfou \and Marc Pickett \and Javier Snaider \and Yun-hsuan Sung \and Brian Strope \and Ray Kurzweil\\
      Google Inc,\\
      1600 Amphitheatre Parkway,\\
      Mountain View, CA 94043, USA\\
      \{{\tt rmyeid, pickett, jsnaider, yhsung, bps, raykurzweil}\}{\tt @google.com}}

\date{}
\begin{document}
\maketitle
\begin{abstract}
We investigate the task of modeling open-domain, multi-turn, unstructured, multi-participant, conversational dialogue.
We specifically study the effect of incorporating different elements of the conversation.
Unlike previous efforts, which focused on modeling messages and responses, we extend the modeling to long context and participant's history.
Our system does not rely on handwritten rules or engineered features; instead, we train deep neural networks on a large conversational dataset.
In particular, we exploit the structure of Reddit comments and posts to extract 2.1 billion messages and 133 million conversations.
We evaluate our models on the task of predicting the next response in a conversation, and we find that modeling both context and participants improves prediction accuracy.
\end{abstract}

\section{Introduction}

Designing conversational systems is a challenging task and one of the original goals of Artificial Intelligence \cite{turing}.
For decades, conversational agent design was dominated by systems that rely on knowledge bases and rule-based mechanisms to understand human inputs and generate reasonable responses \cite{eliza,parry,alice}.
Data-driven approaches emphasize learning directly from corpora of written or spoken conversations.
Recently, this approach gained momentum because of data abundance \cite{serban2015survey}, increasing computational power, and better learning algorithms that automate the feature engineering process \cite{deep1,deep2}.

\begin{table}
\begin{small}
\begin{tabular}{ll}
\textbf{Context 1}: & I live in a village. \\
\textbf{Context 2}: & I live in Chicago. \\
\textbf{Input}: & \emph{Are you going to watch the bulls?} \\
\end{tabular}
\begin{tabular}{lll}
&& \\

\textbf{Response} & \multicolumn{2}{c}{\textbf{Score with}} \\
 &  Context 1 &  Context 2 \\
\hline
I am planning to visit the farm soon. & \textbf{98.35} & 93.91 \\

I am going to watch them on TV. & 94.24 & \textbf{95.35}\\

\end{tabular}
\end{small}
\caption{Our ranker utilizing the context to disambiguate the words \emph{watch} and \emph{bulls} and adjusting the scores of the candidate responses accordingly.}
\label{example}
\end{table}
Here, we study how modeling dialogue is influenced by the history within a conversation, and participants' histories across their conversations.
Recent work in data-driven models focuses on modeling the next response as a function of the preceding message \cite{techstop,diversity}.
We extend previous models in two new directions.
First, we model the history of what has been said before the last message, termed \textbf{\emph{context}}.
This allows the model to include medium-term signals, presumably references and entities, which disambiguate the most recent information.
As the conversation continues and the context grows, we expect our model to make better predictions of the next message (See Table \ref{example}).
Second, to capture longer-term contextual signals, we model each user's personal history across all the conversations in which he or she participated in.
We refer to this information as \textbf{\emph{personal history}}.
The model can personalize its predictions depending on specific users' opinions, interests, experiences, and styles of writing or speaking.
Both of these contextual signals give us the ability to make better predictions regarding future responses.

Characterizing users, language, discourse coherence, and response diversity requires huge datasets and large models.
To gather conversations at scale, we turn to web forums as a source of data.
Specifically, we extract conversations from Reddit, a popular social news networking website.
The website is divided into sub-forums (subreddits), each of which has its own theme of topics and interests.
Registered users can submit URLs or questions, comment on a topic or on other users' comments, and vote on submissions or comments.
Unlike previous efforts that used Twitter as a source of conversations \cite{twitterConv}, Reddit does not have length constraints, allowing more natural text.
We extracted 133 million posts from 326K different subforums, consisting of 2.1 billion comments.
This dataset is several orders of magnitude larger than existing datasets \cite{serban2015survey}.

Instead of modeling message generation directly, the current work focuses on the ranking task of ``\emph{response selection}.''
At each point in the conversation, the task is to pick the correct next message from a pool of random candidates.
Picking the correct next message is likely to be correlated with implicit understanding of the conversation.
We use \textbf{Precision@k} to characterize the accuracy of the system.
We train a deep neural network as a binary classifier to learn the difference between positive, real examples
of input / response pairs, and negative, random examples of input / response pairs.
The classifier's probabilities are used as scores to rank the candidates.
This ranker will choose the response with the highest score.

Unlike generative approaches, where the modeling focus can be dominated by within-sentence language modeling, our
approach encourages the system to discriminate between the connections of an actual response to the current
conversation, and the lack of connections from a random response.
Our model ranks candidates given any subset of the features we discussed so far (i.e., user models, conversation history, or the previous message).
We also jointly learn a shared word embedding space \cite{neuralLM} and a user embedding space.
With this arrangement, the models share common dynamics across users, giving us better models of conversations, and
avoiding the need to construct a different model for each user.

\begin{table}[tb]
\hspace{-1.0em}
{\tiny
\begin{tabular}{rrrrp{0.2\textwidth}}
\textbf{Order} & \textbf{Count} & \multicolumn{1}{c}{\textbf{Avg.\ Score}} & \multicolumn{1}{c}{\textbf{Unique users}} & \multicolumn{1}{c}{\textbf{Comment} }\\ \hline
1&	52259&	1.84&	47537 &	Thanks!\\
2&	50923&	6.85&	43808 &	Yes.\\
4&	35415&	7.14&	31141 &	:( \\  
8&	26976&	2.72&		24169 &	Why? \\
72&	6422&	4.93 & 5838 &	I don't get it. \\
88&	5285&	7.27 &	5085 &	I love you. \\
243&	2559&	2.07&	2482 &	/s\\
267&	2419&	11.10&		2132 &	[deleted]\\
- & 1  & 3 & 1 & Methane hydrates ALSO destroy ozone and there are huge pulses of those when the ice melts it might be that. Do your own research. Xenon fluoride does react with ozone as does the iodine. If you can't find it online try phoning some atmospheric scientists.
\end{tabular}
}
\caption{Reddit comments, in 2014, sorted by their frequency. Frequent comments tend to be short. Users can up-vote and down-vote the score of a comment.}
\label{frequent_comments}

\end{table}

To summarize, our contributions are:
\begin{itemize}[noitemsep,topsep=0pt,parsep=0pt,partopsep=0pt]
\item We model users' responses over long histories that consist of their contributions over the years in various subforums and discussions.
Furthermore, we integrate this model to offer better predictions.
\item We model the conversation history beyond the current message.
We study the length of the modeled history on the performance of our model.
\item We outline a direct path to train and use a discriminative classifier as a response ranker.
\item We demonstrate how to use Reddit's comment structure to extract complex dialogues on a large scale.
We use a scalable neural network architecture that is able to take advantage of the large data size.
\end{itemize}

In Section \ref{related}, we discuss recent relevant work in data-driven modeling of dialogue systems.
Section \ref{reddit} discusses the Reddit dataset's diversity and scale and the steps we took to process the raw data.
In Section \ref{model}, we discuss our choices in conversation modeling with deep neural networks.
We discuss our experimental setup in Section \ref{setup}, analyze our results in Section \ref{results},
and conclude our discussion in Section \ref{conc}.

\section{Related Work}
\label{related}

Ritter et al. \shortcite{twitterConv} proposed a data-driven approach for building dialogue systems,
and they extracted 1.3 million conversations from Twitter with the aim of discovering dialogue acts.
Building on the distributional similarities of the vector space model framework, Banchs and Li \shortcite{iris}
built a search engine to retrieve the most suitable response for any input message.
Other approaches focused on domain specific tasks such as games \cite{games} and restaurants \cite{deeptaskdialog,sds}

Personalizing dialogue systems requires sufficient information from each user and a sufficient user population to define the space.
Writing styles quantified by word length, verb strength, polarity, and distribution of dialogue acts have been used to model users \cite{movie_chars}.
Other efforts focused on building a user profile based on demographics, such as gender, income, age, and marital status \cite{bonin2014context}.
Because Reddit users are pseudo-anonymous, we differ from these approaches by learning the relevant features to model the users' dialogue behavior through embedding each user into a distributed representation.

With the introduction of the sequence-to-sequence framework \cite{seq2seq},
many recent learning systems have used recurrent neural networks (RNNs) to generate novel responses given an input message or sentence.
For example, Vinyals and Le \shortcite{techstop} proposed using IT desk chat logs as a dataset to train LSTM network to generate new sentences.
Sordoni et al. \shortcite{sordoni2015neural} constructed Twitter conversations limiting the history context to one message.
With the help of pre-trained RNN language models, they encoded each message into a vector representation.
To eliminate the need for a language model, Serban et al. \shortcite{serban2015building} tried end-to-end training on an RNN encoder-decoder network.
They also bootstrapped their system with pre-trained word embeddings.

While these systems are able to produce novel responses, it is difficult to understand how much capacity is consumed by modeling
natural language versus modeling discourse and the coherence of the conversation.
Often responses gravitate to the most frequent sentences observed in the training corpus \cite{diversity}.

Perplexity, BLEU, and deltaBLEU, adapted from language modeling and machine translation communities, have been used for evaluating novel responses \cite{yao2015attention,sordoni2015neural,deltableu}.
These metrics only measure the response's lexical fluency and do not penalize for incoherent candidates with regard to the conversational discourse.
While the search for better metrics is still on going, automatic evaluation of response generation stays an open problem \cite{shang2015neural}.

\textbf{Recall@k} or \textbf{Precision@k} are commonly used for measuring a ranker's performance on the task of response selection.
Typically, a positive response is mixed with random responses, and then the system is asked to score the right response higher than others \cite{CNNMatching,kadlec2015improved}.
This task measures the model's ability to discriminate what goes together and what does not.
As these metrics are better understood, we focus on the response selection task in our modeling effort.
\begin{center}
\begin{figure}[tb]
\includegraphics[width=0.5\textwidth]{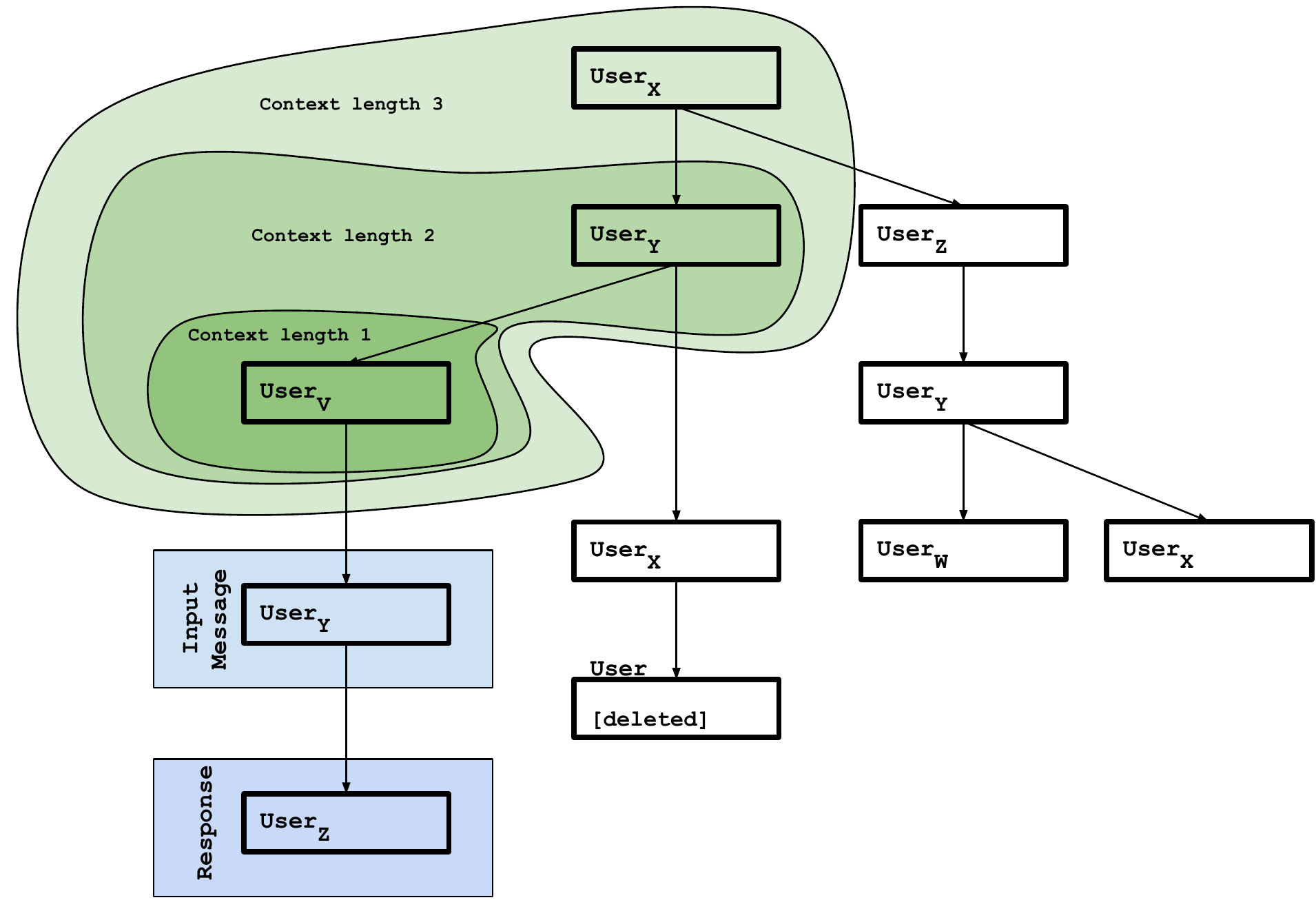}
\caption{A diagram of the Reddit comment tree structure (Reddit Post).
\textbf{$\texttt{User}_\texttt{Z}$} responded to the message produced by \textbf{$\texttt{User}_\texttt{Y}$} (\textcolor{Blue}{\textbf{blue}}). If we follow the ancestors of the input message, we can construct several contexts of different lengths (\textcolor{OliveGreen}{\textbf{green}}).}
\label{reddit_conv}
\end{figure}
\end{center}

\section{Reddit Dataset}
\label{reddit}
As conversational data-driven models are growing in popularity, datasets are increasing in number and size.
However, most are small and domain specific.
Serban et al. \shortcite{serban2015survey} surveyed 56 datasets and found that only 9 have more than 100,000 conversations, only one having more than 5 million conversations.
This limits the complexity and the capacity of the models we can train.
To target open-domain conversations, we need larger datasets.
So far, there has been some limited effort to exploit the rich structure of Reddit.
For example, Schrading et al. \shortcite{reddit_abuse} extracted comments from a small number of subreddits to build a classifier that identifies domestic abuse content.

\begin{figure}[tb]
\hspace*{-3em}
\begin{subfigure}[t]{0.275\textwidth}
\includegraphics[width=\textwidth]{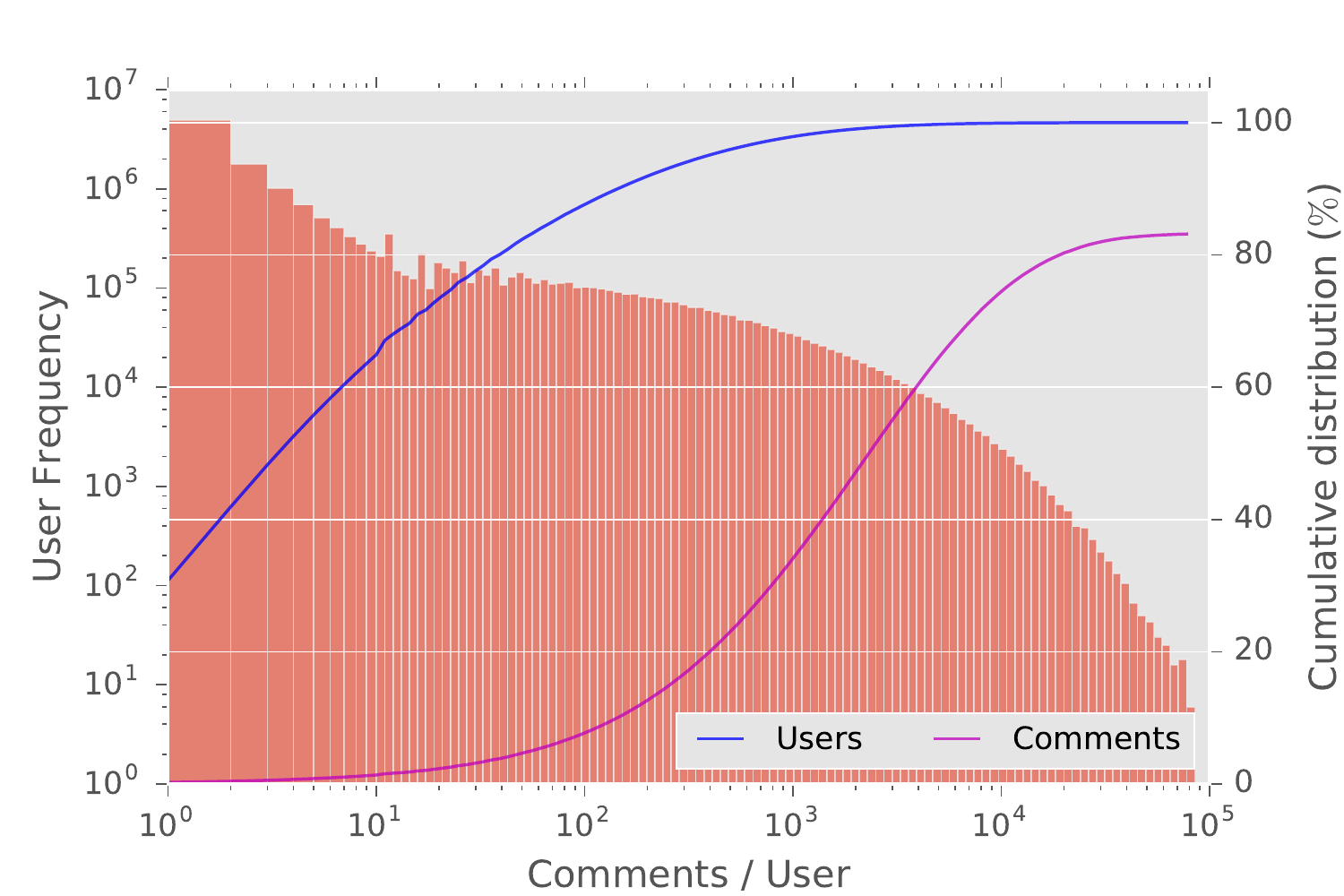}
\caption{Users' comments.}
\label{author_distribution}
\end{subfigure}\hspace*{-0.25em}
\begin{subfigure}[t]{0.275\textwidth}
\includegraphics[width=\textwidth]{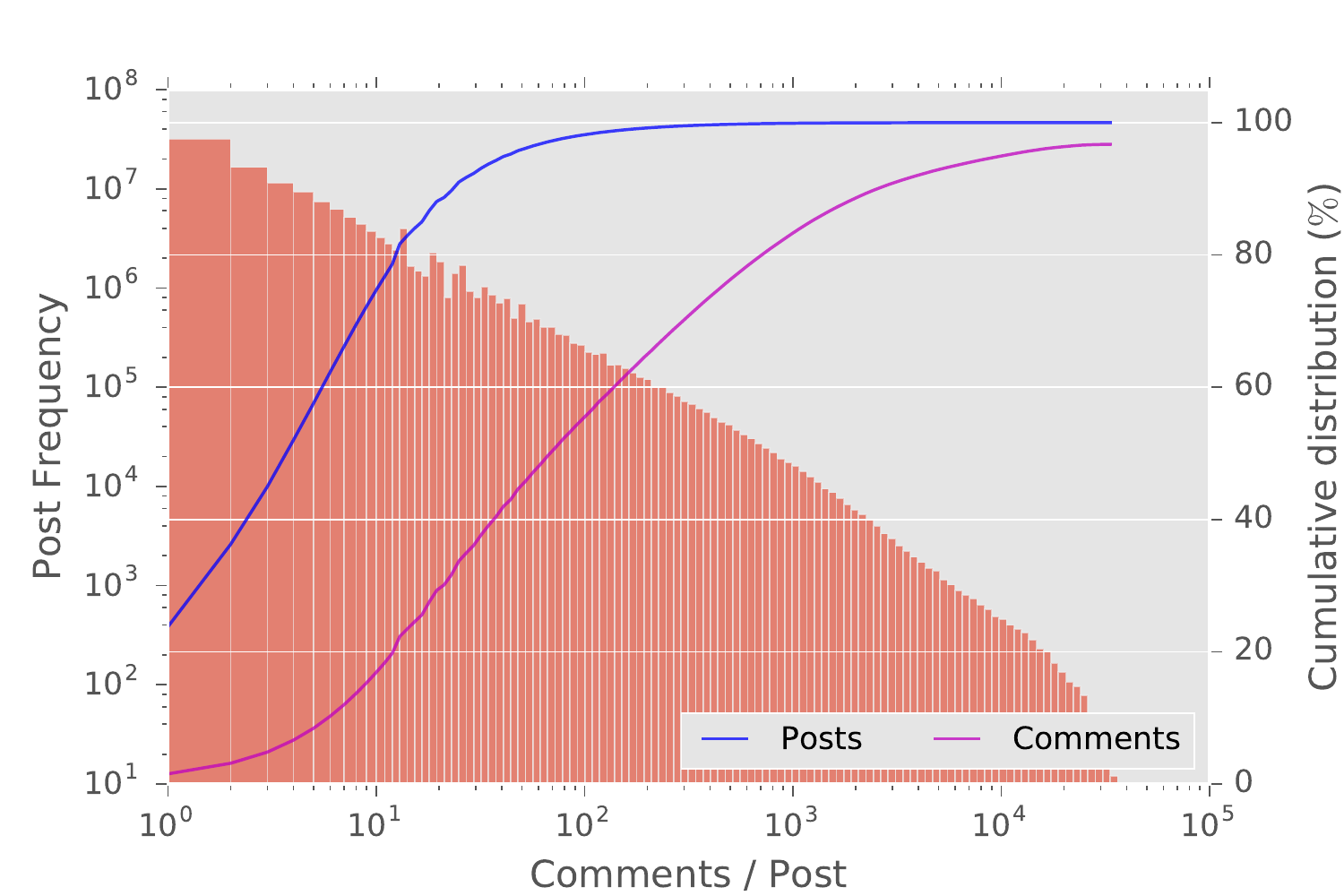}
\caption{Reddit post size.}
\label{post_size}
\end{subfigure}
\begin{subfigure}[t]{0.275\textwidth}
\hspace*{-3em}
\includegraphics[width=\textwidth]{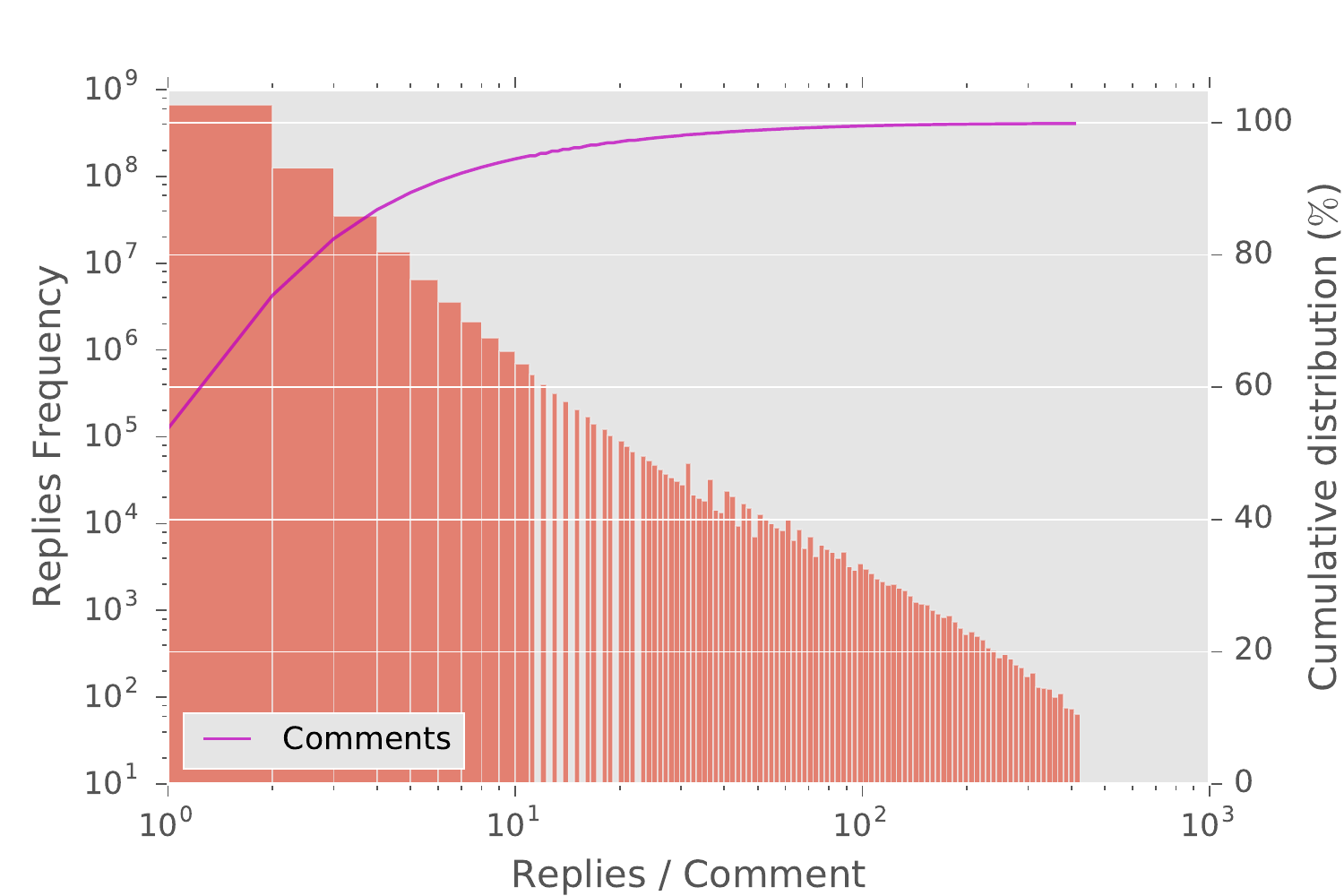}
\caption{Replies per comment.}
\label{responses}
\end{subfigure}\hspace*{-2em}
\begin{subfigure}[t]{0.275\textwidth}
\includegraphics[width=\textwidth]{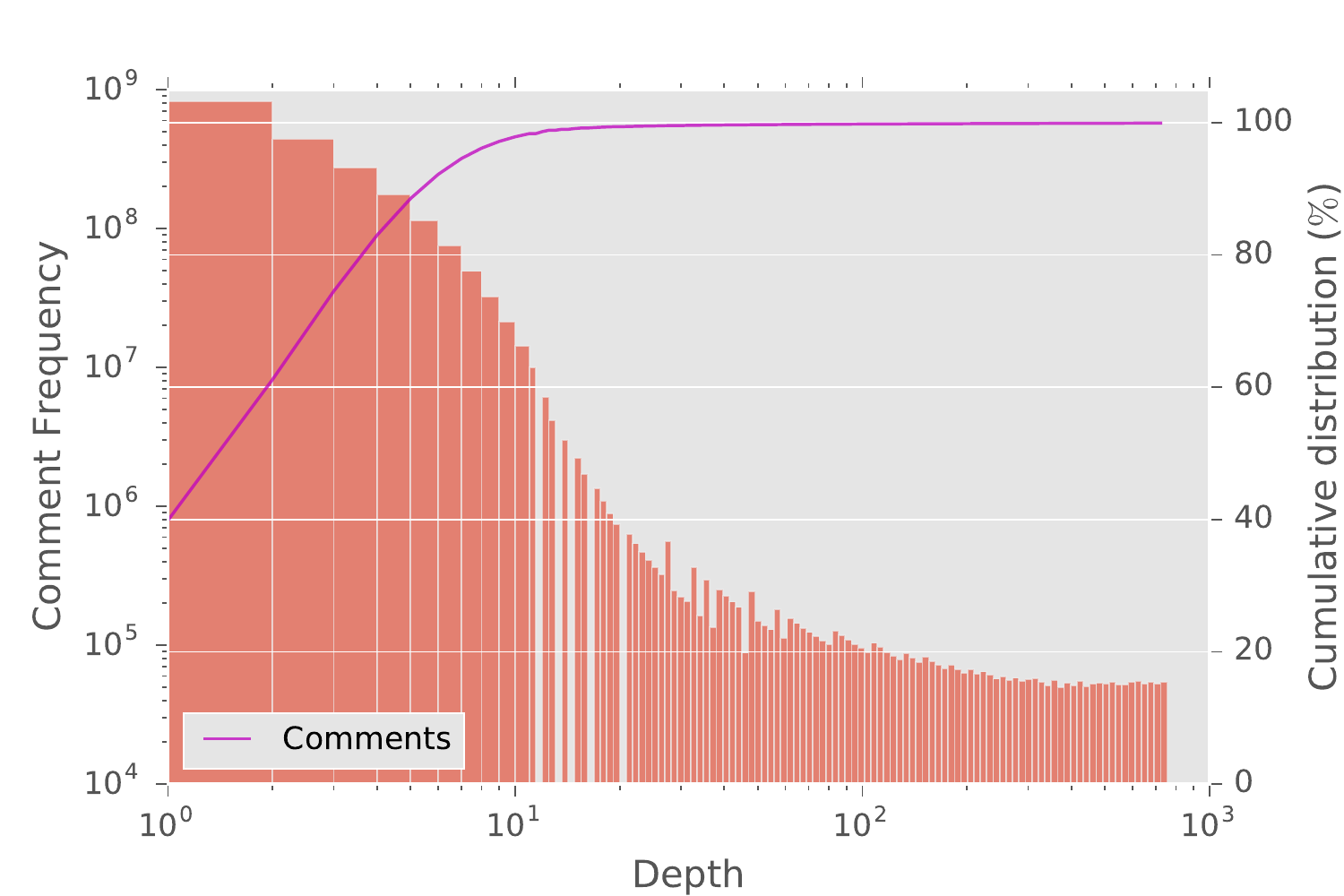}
\caption{Comment depth.}
\label{depth_distribution}
\end{subfigure}
\caption{Reddit dataset statistics. For each metric, we calculate the cumulative distribution of comments that satisfy the criteria.
For example, Figure \ref{post_size} shows that $\sim50\%$ of Reddit posts have less than 100 comments.
}
\label{redditstats}
\end{figure}

Unlike the Ubuntu dataset, logs of technical chat rooms \cite{lowe2015ubuntu}, Reddit conversations tend to be more diverse in regard to topics and user backgrounds.
There are more than 300 thousands sub-forums (subreddits) with different topics of discussion.
Compared to Twitter, Reddit conversations tend to be more natural, as there is no limit on message size (See Table \ref{frequent_comments}).

Figure \ref{reddit_conv} shows how Reddit conversations are organized as a tree or ``Reddit post.''
Users can comment on each other's comments indefinitely, leading to long conversations, which help us construct significant dialogue history.
We construct a conversation by traversing the tree starting at any node, up through its parent and ancestors until we reach the first comment (i.e., the tree's \emph{root}).
Since users cannot comment unless they are registered with a unique user name, we use those names
as our labels for learning our user embedding vectors to personalize the dialogue system.
Note that users tend to be pseudo-anonymous.
They do not use their real names and they participate on the website without sharing private identifying information.

Specifically, we use a public crawl of the reddit website\footnote{\url{ https://bigquery.cloud.google.com/dataset/fh-bigquery:reddit_comments}}.
Figure \ref{redditstats} shows that the dataset is hugely diverse and complex.
Figure \ref{author_distribution} shows that the website has both irregular contributors and heavy users who have a large number of comments.
Unlike the datasets surveyed in \cite{serban2015survey}, a comment can have several user generated responses.
While these diverse responses by no means cover the space of reasonable responses, this property may help our models in generalization (See Figure \ref{responses}).

\section{Models}
\label{model}
We define a conversation $\mathcal{C}$ to be a sequence of $k$ pairs of \underline{M}essages and participants (\underline{A}uthors) $\mathcal{C} \equiv \left((M_1, A_1), (M_2, A_2), \dots, (M_k, A_k)\right)$.
Here, a message $M_i$ is a sequence of a variable number of words $M_i = (w_{i1}, w_{i2}, .... w_{il})$.
$A_i$ and $w_j$ are random variables taking values in the user population $\mathcal{P}_{user}$ and the word vocabulary $\mathcal{V}_{word}$, respectively.
$\mathcal{P}_{user}$ is a fixed set of the most frequent authors in reddit, it is basically, a dictionary of their usernames that is used to index the author embedding matrix.
Every vector is limited to only one user.

To represent messages we use bag of words technique over recurrent or convolutional networks for its speed and ability to scale to a dataset as large as Reddit.
To improve bag of words capability of keeping track of words' order and sentence structure, we define $\mathcal{V}_{ngram}$ to be a dictionary of a subset of ngrams defined over the word vocabulary.

The first step in our modeling is to map each user in our population $\mathcal{P}_{user}$ and each word  in our vocabulary $\mathcal{V}_{word}$ to a vector of $d$ dimensions.
Specifically, we define $\phi_{user}: A_i \mapsto \mathbb{R}^{d_A}$ to be the embedding of the user $A_i$ and $\phi_{ngram}: (w_i, w_j, \dots) \mapsto \mathbb{R}^{d_n}$ to be the embedding of the ngram $(w_i, w_j, \dots)$.
For a sequence of $k$ messages, we define the bag of ngrams embedding ($\psi \in \mathbb{R}^{d_n}$) to be the average of the embeddings of the ngrams extracted from all the messages:
\begin{equation}
\psi({M_1, \dots, M_k}) = \frac{1}{L}\sum_{1 \le j \le k}\ \  \sum_{g \in ngrams(M_j)}{\phi_{{ngram}}(g)}
\end{equation}
where $L$ is the total number of ngrams extracted from all the messages $\{M_1, \dots, M_k\}$.
Next, for a sequence of message-participant pairs of length $k$, we define the following features:
\begin{itemize}[noitemsep,topsep=0pt,parsep=0pt,partopsep=0pt]
\item \textbf{Response} : $\mathbf{R} = \psi(M_k)$ where $M_{k}$ is the last message in the sequence.
\item \textbf{Input Message}: $\mathbf{I} = \psi(M_{k-1})$ where $M_{k-1}$ is the message that the response is addressing.
\item \textbf{Context}: $\mathbf{C} = \psi(M_1, M_2, \dots M_{k-2})$ where $(M_1, M_2, \dots M_{k-2})$ is the subsequence of messages that preceded the input message.
\item \textbf{Author}: $\mathbf{A} = \phi_{user}(A_k)$ where $A_k$ is the user who generated the response message.
\end{itemize}

\begin{figure}[tb]
\includegraphics[width=0.5\textwidth, trim={0 0 0 0}, clip]{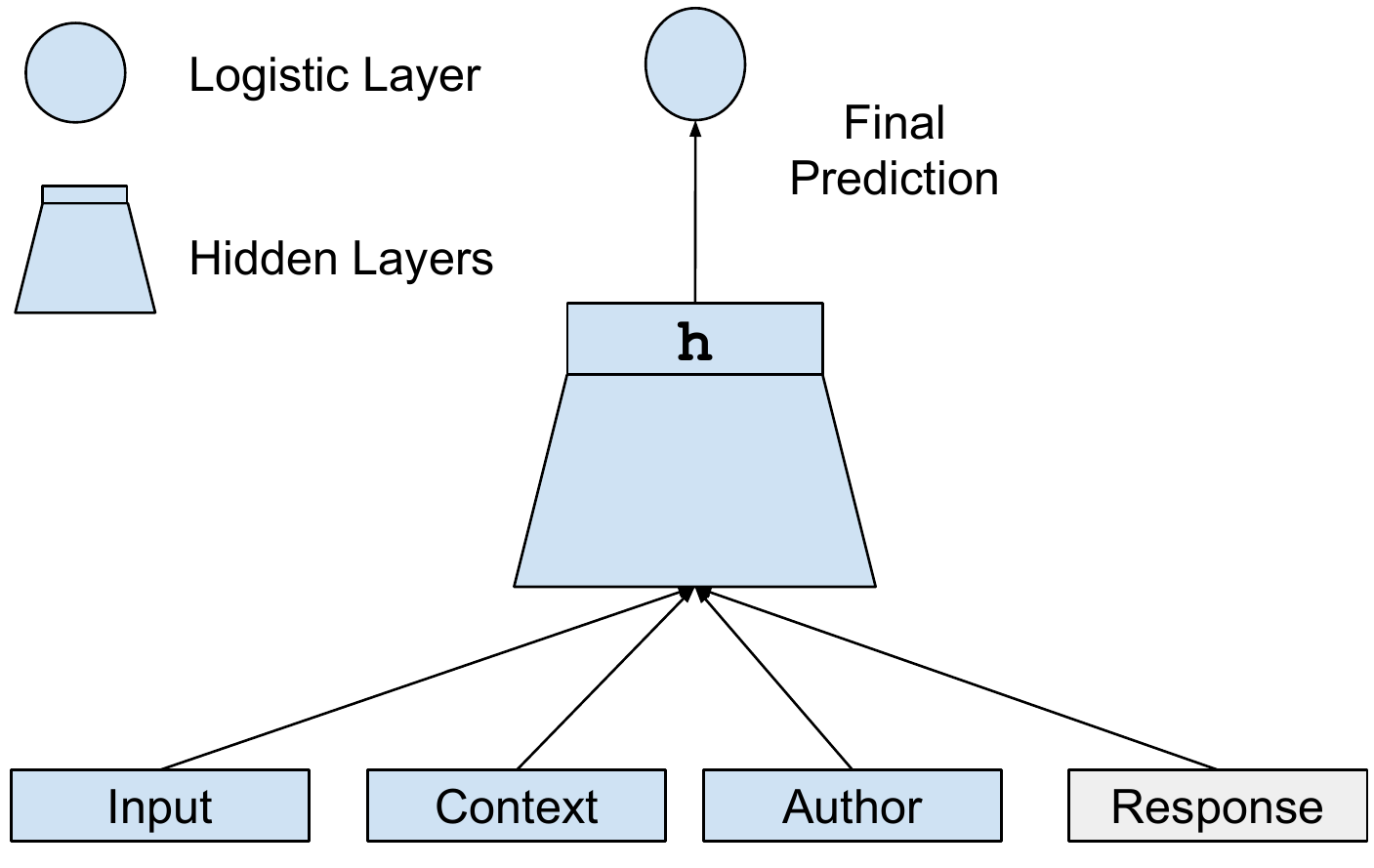}
\caption{Single loss model.}
\label{small_model}
\end{figure}

In Reddit, for each message in the post tree, we consider its parent to be the input message and its parent's ancestors to be the context.
The content of the message is the response and the user that wrote the message is the author.

\subsection{Response Ranking}
To measure the effect of our features on modeling conversations, our task is to select the best response out of a pool of random candidates.
This selection process could be viewed as a ranking problem.
There are several approaches to ranking: pointwise, pairwise, and list-wise \cite{ranking}.
Kadlec et al. \shortcite{kadlec2015improved} chose pointwise ranking for its simplicity, and we follow the same choice for its speed benefits, which are necessary for training on hundreds of millions of examples.
In pointwise ranking, we consider the compatibility of only one candidate at a time.
Specifically, we learn a model that estimates the probability of a candidate given a subset of the features $\{\mathbf{I}, \mathbf{C}, \mathbf{A}\}$.

To construct the training dataset, we form pairs of features and responses.
For each response appearing in the corpus, we form two pairs.
The first is composed of the features with the observed response $(\{\mathbf{I}, \mathbf{C}, \mathbf{A}\}, \mathbf{R})$.
In the second pair, we replace the response with another random response sampled from our corpus, $(\{\mathbf{I}, \mathbf{C}, \mathbf{A}\}, \mathbf{R'})$.
The first pair is used as a positive example and the second is a negative one.

\begin{figure}[t]
\includegraphics[width=0.5\textwidth, trim={0 0 0 0}, clip]{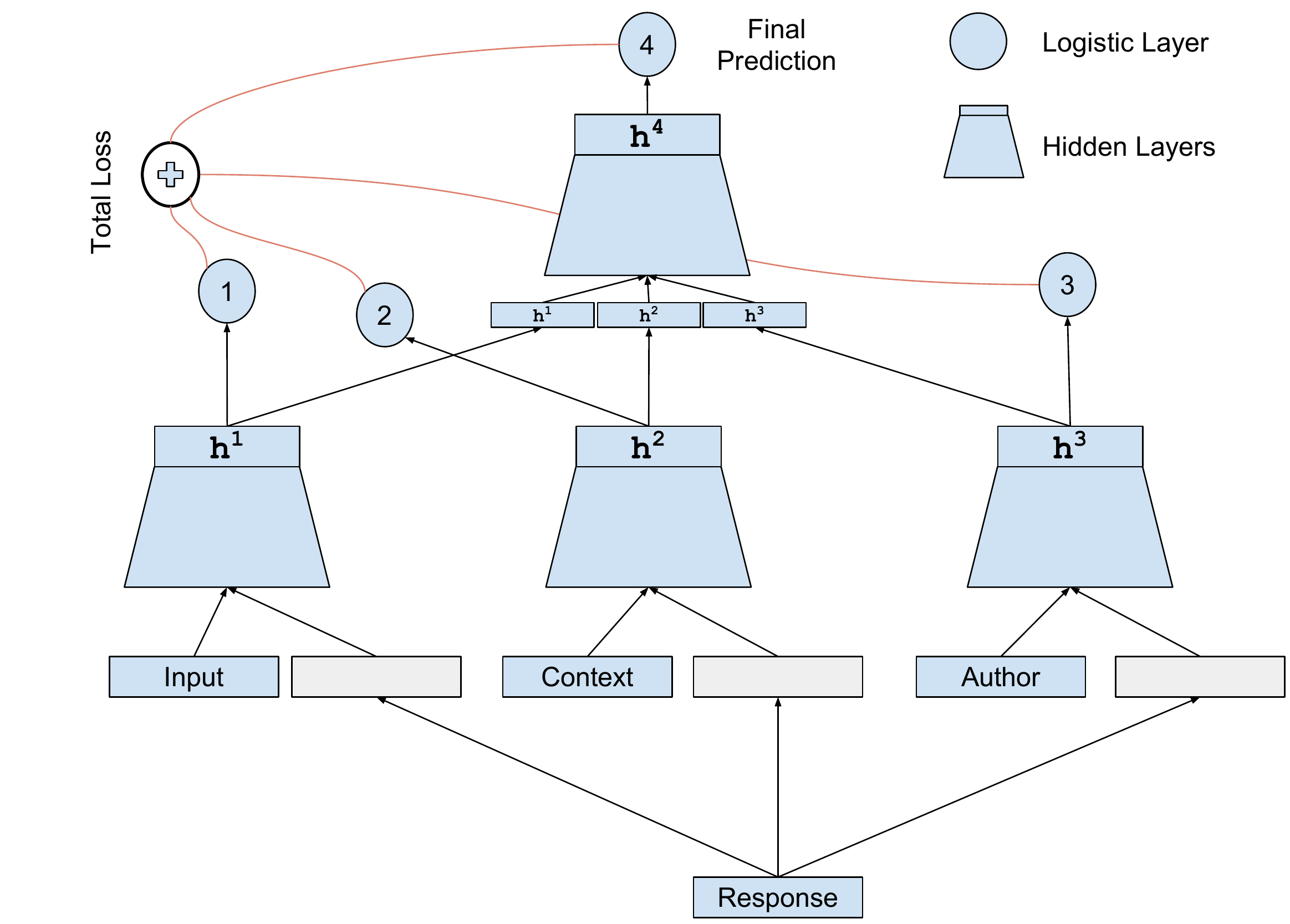}
\caption{Multi-loss model.}
\label{large_model}
\end{figure}

\subsection{Single-loss Network}
To estimate the probability of the response appearing given the features, we train a binary logistic regression classifier.
Figure \ref{small_model} shows a network that concatenates the previous features into one input vector $input = [\mathbf{I}; \mathbf{C}; \mathbf{A}; \mathbf{R}]$.
Then, several hidden layers with Relu non-linearities follow to produce a hidden layer $\mathbf{h}$.
Given the hidden layer $\mathbf{h}$, we estimate the probability of the response, as follows:
\begin{equation}
\Pr(\mathbf{R}| \mathbf{I}, \mathbf{C}, \mathbf{A}) \approx \sigma(\mathbf{W}\mathbf{h} + \mathbf{b})
\end{equation}
Where $\sigma$ is the sigmoid function $\sigma(x) = 1/(1+e^{-x})$.
We call this model a single-loss model because it makes one prediction using all the information available.

\subsection{Multi-loss Network}

We can formalize the previous single-loss model network further by declaring
$\Pr(\mathbf{R}|\mathbf{x}) \approx Network(\mathbf{x})$, where $\mathbf{x}$ is the input feature vector.
The network uses a logistic regression layer on top of a feed-forward neural network.
Figure \ref{large_model} shows the multi-loss architecture that could be viewed as a network of networks.
This architecture is achieved by replicating the single-loss architecture ($Network$) three times for each feature.
Each of the networks makes predictions taking into account one feature at a time.
Furthermore, each network produces a hidden layer ($\mathbf{h^i}$) that will be used in an aggregate network.
The aggregate network concatenates the hidden layers from the previous networks, $\mathbf{[h^1; h^2; h^3]}$, to produce a final hidden layer $\mathbf{h^4}$.
This allows the final prediction to take advantage of all the features jointly.
This network also allows us to measure the performance of each feature alone.
This modular architecture facilitates diagnosis of any possible problems during training.

More specifically, the networks represent the following classification problems:
\begin{align*} 
\Pr(\mathbf{R}| \mathbf{I})  & \approx  \sigma(\mathbf{W_1}\mathbf{h^1} + \mathbf{b_1}) \\
\Pr(\mathbf{R}| \mathbf{C}) & \approx \sigma(\mathbf{W_2}\mathbf{h^2} + \mathbf{b_2}) \\
\Pr(\mathbf{R}| \mathbf{A}) & \approx \sigma(\mathbf{W_3}\mathbf{h^3} + \mathbf{b_3}) \\
 \Pr(\mathbf{R}| \mathbf{I}, \mathbf{C}, \mathbf{A}) & \approx  \sigma(\mathbf{W_4}\mathbf{h^4} + \mathbf{b_4})
\end{align*} 

We use only the final prediction in the evaluation $\Pr(\mathbf{R}| \mathbf{A}, \mathbf{C}, \mathbf{I})$, but we penalize the model with the sum of all predictions' losses.

\section{Experimental Setup}
\label{setup}
We extract 2.1 Billion comments that were posted on the Reddit website between 2007 and 2015.
We group the comments by their post page, treating each Reddit post as a tree rooted at the title of the post.
We generate a positive example from each comment in the post.
The example features are calculated by looking at the message's attributes, its parent, and its ancestors in the tree.
We exclude Reddit posts that have more than 1000 comments for computational reasons.
Most of these large posts are ``Mega-threads", each containing hundreds of thousands of comments.
We do not generate examples for comments with empty or missing input message features.
We also exclude examples where the author is not in our user population $\mathcal{P}_{user}$, or the user profile was deleted.
After this filtering, 550 million positive examples are yielded.
For each positive example, we generate a negative example by replacing the response feature by a random comment from Reddit.

\subsection{Vocabulary}
Reddit comments are written in markdown markup language.
First, we remove the markdown and then, tokenize the textual content.
We normalize URLs and then include in our vocabulary the most frequent 200K unigrams and 200K bigrams.
The count of the least frequent unigram is $1229$, and for the least frequent bigram is $27670$.
For the author embeddings, we construct a user population ($\mathcal{P}_user$) of the most frequent contributing 400K users.
The least contributing user created $922$ comments.
This population is, essentially, a dictionary of the user names.

\subsection{Training}
We set the ngram embedding and the user embedding space to 300 dimensions.
The single loss model consists of one network, while the multiloss network consists of four networks.
Each network consists of three hidden layers of size $[500, 300, 100]$.
The hidden layer parameters, the ngram embeddings, and the user embeddings are trained jointly, and we use stochastic gradient descent (SGD) to optimize the parameters of our models \cite{sgd}.
The derivatives are estimated using the back-propagation algorithm and the updates are applied asynchronously.
The learning rate $\alpha$ for SGD is set to $0.03$.
The models are implemented using TensorFlow \cite{tensorflow}.
Despite that each model is trained on 5 GPUs, the training time still takes several weeks due to data size.

We split our dataset into three partitions: train, dev, and test.
The training dataset consists of $90\%$ of the data and the rest is divided between dev and test.
We train our classifier for one epoch, which is equal to 1 Billion examples.
We stop training our models when we observe no significant improvement in the accuracy of our binary classifier on the dev dataset.

As our binary classifier will be evaluated on ranking candidate responses, we extract 10,000 examples from our test dataset.
For each example, we sample $N-1$ random responses from the pool.
Given $N$ of candidates, the classifier is asked to give the highest probability to the positive candidate available in the pool.
We report precision (\textbf{P@1}) as a metric of quality.

\section{Discussion \& Results}
\label{results}

In this section, we discuss the gains achieved by integrating the conversation history as well as the participants' history into our modeling.
We contrast both approaches and contrast their qualities and show a final model that takes advantage of both.
Then, we show the effect of increasing the training dataset size on our models performance.

\subsection{Length of the Context}

How far back do we need to look to improve the quality of our ranking?
To test that, we train both models discussed in Section \ref{model} on several datasets with context features that vary in temporal scope.
Table \ref{history_length_effect} shows the Precision @ 1 for both models using two different ranking tasks,  the first involves 10 candidates and the second has 100 candidates.
Context of length 0 corresponds to using only the input message as a feature.
Each model was trained and tested on examples that included a conversation history (\textbf{context length}) \emph{up to} $m$ number of messages and not necessarily all the messages in the training or the test included the same history length.
\begin{table}[tb]
\hspace*{-1.0em}
\centering
\begin{tabular}{ccccc}
\multicolumn{1}{l}{}                         & \multicolumn{4}{c}{Model}                                                          \\
\textbf{}                                    & \multicolumn{2}{c}{\textbf{Single-loss}} & \multicolumn{2}{c}{\textbf{Multi-loss}} \\
\multicolumn{1}{l|}{Context Length}                        & \multicolumn{2}{c|}{\textbf{$N$}}   & \multicolumn{2}{c|}{\textbf{$N$}}  \\
\multicolumn{1}{c|}{Up To} &  10   & \multicolumn{1}{c|}{100}  &  10  & \multicolumn{1}{c|}{100}  \\ \hline
\multicolumn{1}{c|}{0}                                         &      74.45    & \multicolumn{1}{c|}{47.92}         &     75.15    &  \multicolumn{1}{c|}{49.16}         \\
\multicolumn{1}{c|}{1}                       &     78.38     & \multicolumn{1}{c|}{51.80}         &   80.16      & \multicolumn{1}{c|}{55.97}         \\
\multicolumn{1}{c|}{2}                       &     79.23     & \multicolumn{1}{c|}{52.30}         &   81.30      & \multicolumn{1}{c|}{56.25}         \\
\multicolumn{1}{c|}{5}                       &  78.41         & \multicolumn{1}{c|}{\textbf{52.84}}         &       81.35  & \multicolumn{1}{c|}{\textbf{56.39}}         \\
\multicolumn{1}{c|}{10}                       &  79.32         & \multicolumn{1}{c|}{50.74}         &      81.70  & \multicolumn{1}{c|}{55.25}         \\
\multicolumn{1}{c|}{25}                      &   \textbf{79.70}       & \multicolumn{1}{c|}{51.70}         &    \textbf{81.71}     & \multicolumn{1}{c|}{55.52}         \\

\end{tabular}
\caption{Precision @ 1 for models trained on different context lengths and tested on two different sizes of candidates pools.}
\label{history_length_effect}
\end{table}

First, we observe clear gains when we integrate the context feature ($\mathbf{C}$).
\textbf{P@1} increases by 4-6 points the moment we include the message that preceded the input message.
However, we see a diminishing return as the context increases, particularly when the context is larger than $5$ messages.
In this case, there could be two factors at work.
First, the more messages we use, the larger the number of vectors we average;
this tends to blur the features and increase the information loss compared to the insight we gain.
Second, we have less training and a smaller number of test examples that could take advantage of a long history.
Figure \ref{depth_distribution} shows that more than $90\%$ of the reddit comments haver a lower depth than 6 messages in the tree.

\subsection{Personalization}
Table \ref{author_contribution} shows larger gains in our rankers' precision when using the author feature compared to the conversational history (context) feature.
The multi-loss model improves by $5$ points in the task of ranking $100$ response candidates.
The author vector represents longer historical information than the current conversation history.
Personal history could include interests, opinions, demographics, writing style, and personality traits.
These could be essential in determining if a response is appropriate.

\begin{table}[tb]
\hspace*{-1.0em}
\begin{tabular}{ccccc}
\multicolumn{1}{l}{}                         & \multicolumn{4}{c}{Model}                                                          \\
\textbf{}                                    & \multicolumn{2}{c}{\textbf{Single-loss}} & \multicolumn{2}{c}{\textbf{Multi-loss}} \\
\multicolumn{1}{l|}{}                        & \multicolumn{2}{c|}{\textbf{$N$}}   & \multicolumn{2}{c|}{\textbf{$N$}}  \\
\multicolumn{1}{c|}{\textbf{Feature}} &  10   & \multicolumn{1}{c|}{100}  &  10  & \multicolumn{1}{c|}{100}  \\ \hline

\multicolumn{1}{c|}{message}                       &      74.45    & \multicolumn{1}{c|}{47.92}         &     75.15    &  \multicolumn{1}{c|}{49.16}         \\

\multicolumn{1}{c|}{message + context}                      &   79.70       & \multicolumn{1}{c|}{51.70}         &    81.71     & \multicolumn{1}{c|}{55.52}         \\


\multicolumn{1}{c|}{message + author}                       &     79.52     & \multicolumn{1}{c|}{53.03}         &   83.25      & \multicolumn{1}{c|}{60.53}         \\

\multicolumn{1}{c|}{All}                      &    \textbf{82.72}     & \multicolumn{1}{c|}{\textbf{55.91}}         &   \textbf{86.60}      & \multicolumn{1}{c|}{\textbf{63.53}}         \\

\end{tabular}
\caption{\textbf{P@1} improvement gained by adding author and/or context to the base model.
We consider context length to be 25.}
\label{author_contribution}
\end{table}

Finally, if we use all the features available to us, we get further improvement in performance over any of the features used alone.
This highlights that the information we recover from each feature is different.

\subsection{Multi-loss Vs Single-loss}

The motivation behind the multi-loss model is to prevent adaptation between features \cite{coadaptation}.
In the single-loss model, the author feature could be subsumed for many cases with the input message and the context.
Only subtle cases may require knowing the author identity to determine if the response is suitable.
This slows the learning process of good author vectors.
Therefore, the multi-loss network requires that the author vector should capture enough information to perform the prediction task, solely.
This architecture extends the \emph{deep supervision} idea where companion objective function is introduced to train intermediate layers in a deep network \cite{deepsupervision}.
Notice how the author feature outperforms the context feature in all tasks with the introduction of the multi-loss model.
The multi-loss model is also easier to debug and probe.
By reporting every loss on its own, we can see the development of the network.

\begin{figure}[tb]
\vspace{-1.25em}
\includegraphics[width=0.5\textwidth]{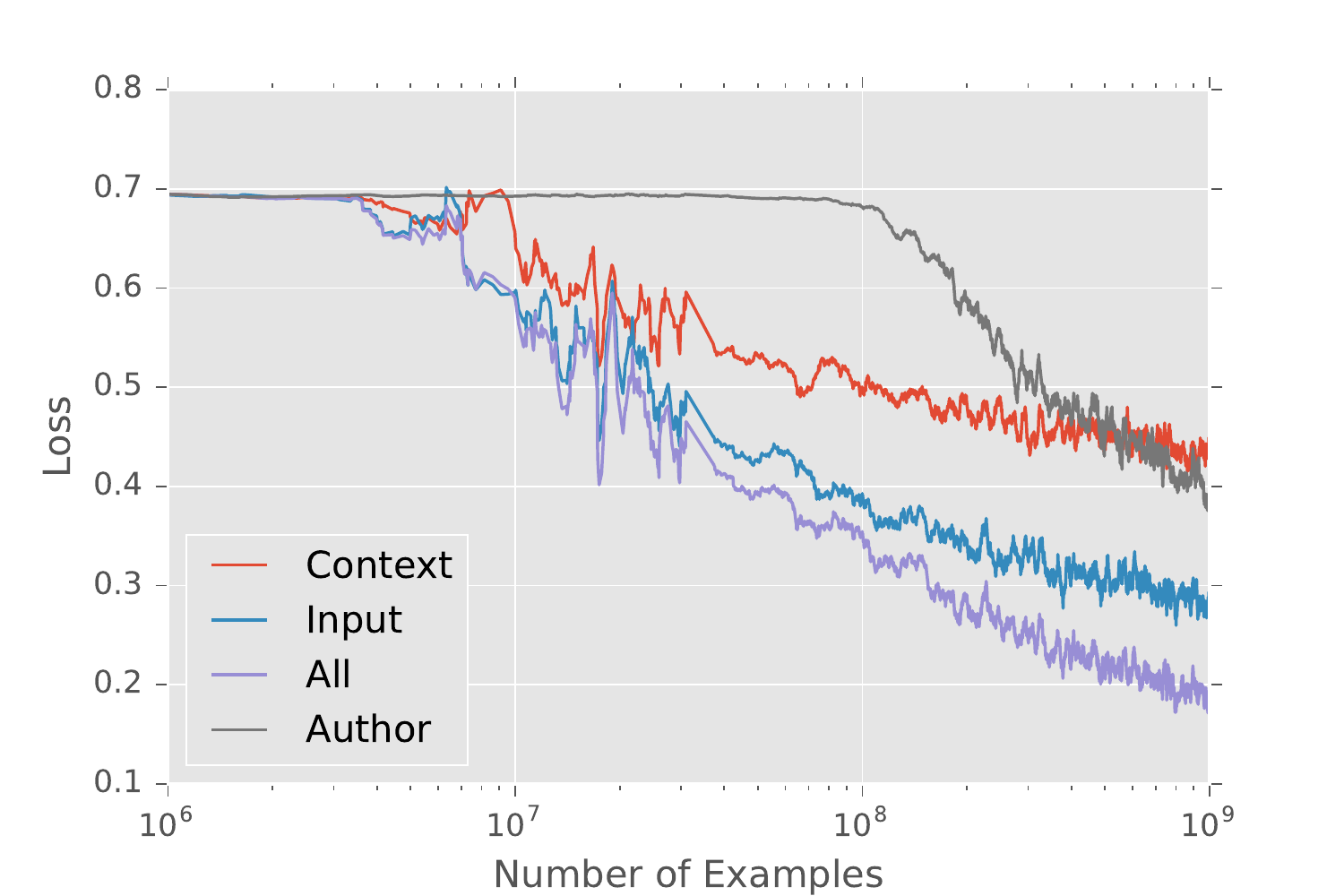}
\caption{Losses contributed by each feature in the multi-loss network. Combining all the features always produce lower loss.}
\label{losses}
\end{figure}

Figure \ref{losses} shows the loss contributed by each feature.
Notice, how the author vector takes more than 100 million examples to start influencing the prediction task.
We conjecture that this behavior is the result of two factors.
First, the author distribution does not follow Zipf's law, as language does.
There is no small number of authors that could cover most of the examples.
Second, author vectors depend. indirectly, on the content of the comments they posted.
Unless the representation of the language, and consequently messages, are stable, we cannot learn a aggregate representation of the user's set of messages.
This multi-stage learning is similar to what McClelland and Rogers \shortcite{mcclelland} observed in their work.

\subsection{New users}
In our evaluation we did not consider the case of unknown users.
However, if a new user is encountered by our model, we can add a randomly initialized vector as a temporary representation.
As the conversation goes on, we can then refine user vector using backpropagation while the rest of the model
parameters are fixed.
This technique is similar to the paragraph vector's strategy of dealing with new paragraphs after training is finished \cite{paragraph_vector}.

\begin{figure}[t]
\vspace{-.4em}
\begin{subfigure}[]{0.25\textwidth}
\includegraphics[width=\textwidth]{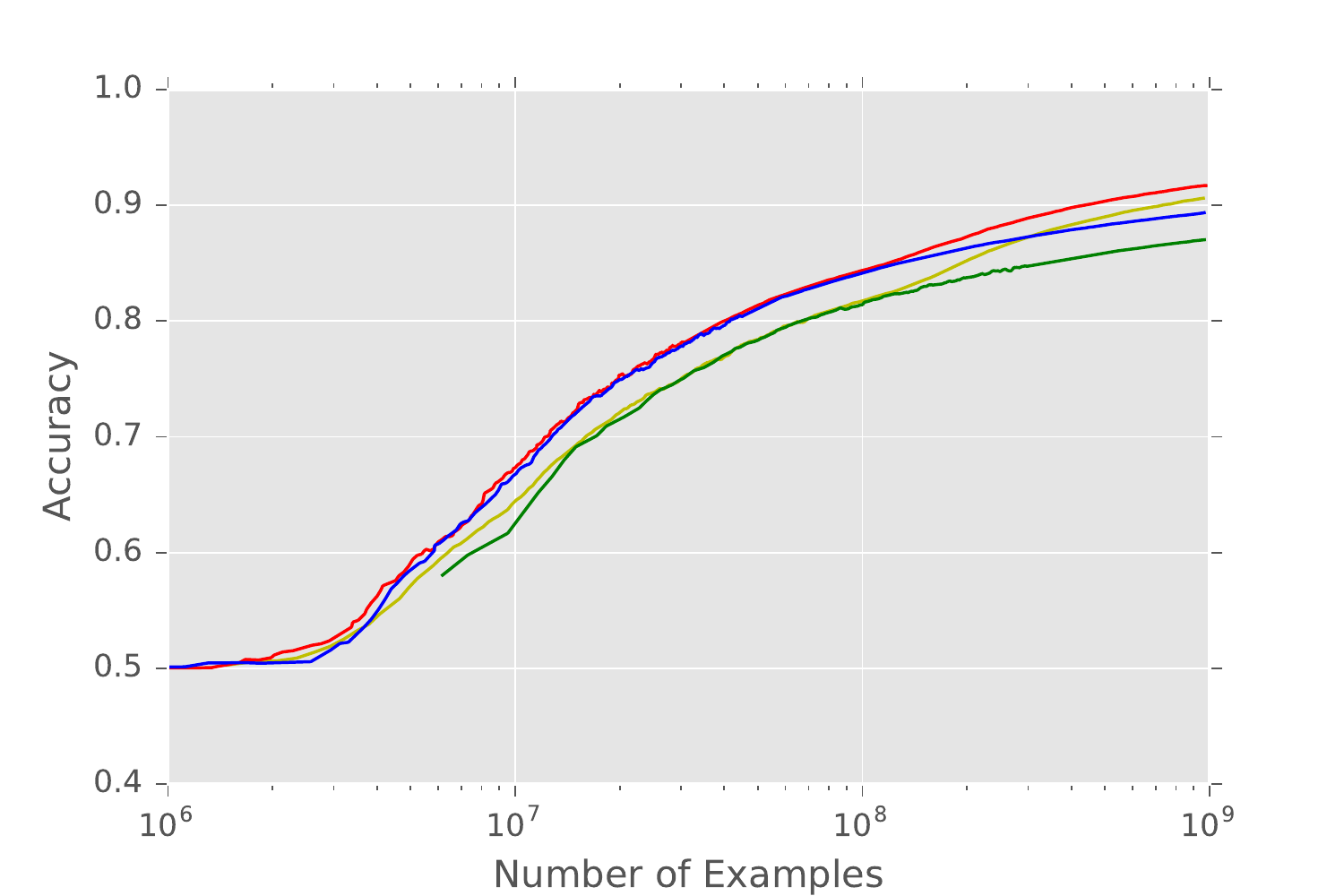}
\caption{Learning curve.\vspace{1.25em}}
\label{learning_curve}
\end{subfigure}\hspace*{-0.25em}
\begin{subfigure}[]{0.25\textwidth}
\includegraphics[width=\textwidth]{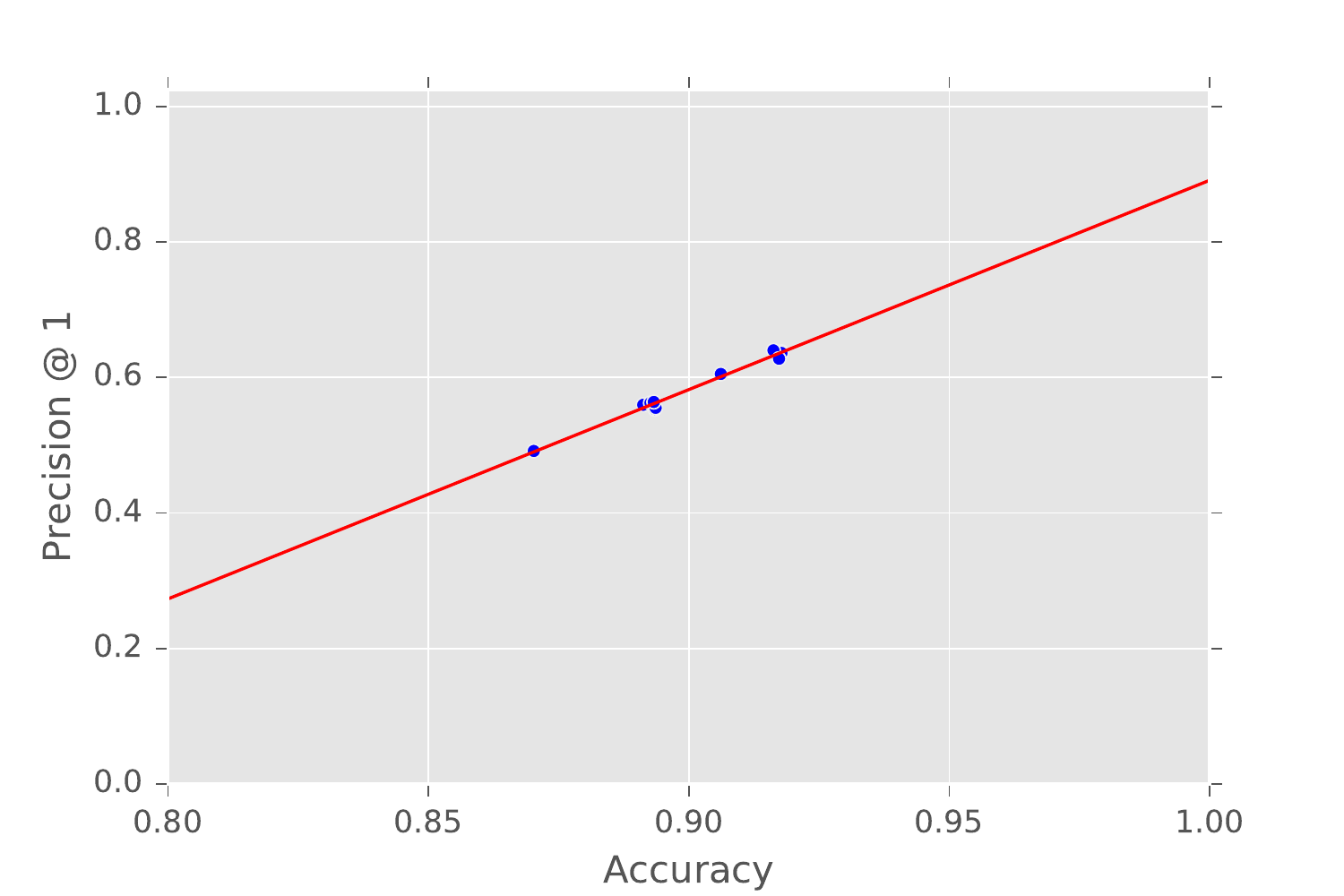}
\caption{Classifier Accuracy Vs. Ranking Quality.}
\label{future_precision}
\end{subfigure}
\caption{Effect of the training dataset size on the binary classifier accuracy, and therefore, the ranker precision. We used the multi-loss model with all the features and $N$ set to $100$.}
\label{correlation}
\end{figure}

\subsection{Learning Curves}
Figure \ref{learning_curve} shows the improvement of the classifier accuracy by increasing the training dataset size orders of magnitude at a time.
The results we presented so far would not have been possible without the billion examples we extracted from Reddit.
It is quite clear that our models would have performed poorly given the other previously used datasets given their small sizes.

Moreover, the accuracy of the binary classifier is correlated highly with the \textbf{P@1} of the rankers we evaluated.
We found that the pearson correlation between accuracy observed on the dev dataset and \textbf{P@1} of the ranker tested on the test dataset is both strong and positive, between $+0.94$ and $+0.99$.
Therefore, we may infer the future gains of increasing the size of the dataset on the quality of the ranker (See Figure \ref{future_precision}).

\section{Conclusion}
\label{conc}
Using Reddit, our model is trained on a significantly larger conversational dataset than previously published efforts.
We train two scalable neural network models using bags of ngram embeddings and user embeddings.
We measure significant improvement in the task of selecting the next response by integrating what has been said in the conversation so far.
We study the effect of the length of the conversation history on performance.
We also personalize the selection process by learning an identity feature for each user.
This yields further improvement as it models the longer history of what a user has said in all conversations.
Finally, our multi-loss model shows improvements over the baseline single-loss model using any subset of the features.

\bibliography{references}

\begin{thebibliography}{}

\bibitem[\protect\citename{Abadi \bgroup et al.\egroup }2015]{tensorflow}
Mart{\i}n Abadi, Ashish Agarwal, Paul Barham, Eugene Brevdo, Zhifeng Chen,
  Craig Citro, Greg~S Corrado, Andy Davis, Jeffrey Dean, Matthieu Devin, et~al.
\newblock 2015.
\newblock Tensorflow: Large-scale machine learning on heterogeneous systems,
  2015.
\newblock {\em Software available from tensorflow. org}.

\bibitem[\protect\citename{Banchs and Li}2012]{iris}
Rafael~E. Banchs and Haizhou Li.
\newblock 2012.
\newblock Iris: a chat-oriented dialogue system based on the vector space
  model.
\newblock In {\em Proceedings of the ACL 2012 System Demonstrations}, pages
  37--42, Jeju Island, Korea, July. Association for Computational Linguistics.

\bibitem[\protect\citename{Bengio \bgroup et al.\egroup }2006]{neuralLM}
Yoshua Bengio, Holger Schwenk, Jean-S{\'e}bastien Sen{\'e}cal, Fr{\'e}deric
  Morin, and Jean-Luc Gauvain.
\newblock 2006.
\newblock Neural probabilistic language models.
\newblock In {\em Innovations in Machine Learning}, pages 137--186. Springer.

\bibitem[\protect\citename{Bonin \bgroup et al.\egroup }2014]{bonin2014context}
Francesca Bonin, Jose San~Pedro, and Nuria Oliver.
\newblock 2014.
\newblock A context-aware nlp approach for noteworthiness detection in
  cellphone conversations.
\newblock In {\em COLING}, pages 25--36.

\bibitem[\protect\citename{Bottou}1991]{sgd}
{L\'eon} Bottou.
\newblock 1991.
\newblock Stochastic gradient learning in neural networks.
\newblock In {\em Proceedings of Neuro-N\^imes 91}, Nimes, France. EC2.

\bibitem[\protect\citename{Cuay{\'{a}}huitl}2016]{sds}
Heriberto Cuay{\'{a}}huitl.
\newblock 2016.
\newblock Simpleds: {A} simple deep reinforcement learning dialogue system.
\newblock {\em CoRR}, abs/1601.04574.

\bibitem[\protect\citename{Galley \bgroup et al.\egroup }2015]{deltableu}
Michel Galley, Chris Brockett, Alessandro Sordoni, Yangfeng Ji, Michael Auli,
  Chris Quirk, Margaret Mitchell, Jianfeng Gao, and Bill Dolan.
\newblock 2015.
\newblock deltableu: A discriminative metric for generation tasks with
  intrinsically diverse targets.
\newblock In {\em Proceedings of the 53rd Annual Meeting of the Association for
  Computational Linguistics and the 7th International Joint Conference on
  Natural Language Processing (Volume 2: Short Papers)}, pages 445--450,
  Beijing, China, July. Association for Computational Linguistics.

\bibitem[\protect\citename{Hinton \bgroup et al.\egroup }2012]{coadaptation}
Geoffrey~E Hinton, Nitish Srivastava, Alex Krizhevsky, Ilya Sutskever, and
  Ruslan~R Salakhutdinov.
\newblock 2012.
\newblock Improving neural networks by preventing co-adaptation of feature
  detectors.
\newblock {\em arXiv preprint arXiv:1207.0580}.

\bibitem[\protect\citename{Hu \bgroup et al.\egroup }2014]{CNNMatching}
Baotian Hu, Zhengdong Lu, Hang Li, and Qingcai Chen.
\newblock 2014.
\newblock Convolutional neural network architectures for matching natural
  language sentences.
\newblock In Z.~Ghahramani, M.~Welling, C.~Cortes, N.D. Lawrence, and K.Q.
  Weinberger, editors, {\em Advances in Neural Information Processing Systems
  27}, pages 2042--2050. Curran Associates, Inc.

\bibitem[\protect\citename{Kadlec \bgroup et al.\egroup
  }2015]{kadlec2015improved}
Rudolf Kadlec, Martin Schmid, and Jan Kleindienst.
\newblock 2015.
\newblock Improved deep learning baselines for ubuntu corpus dialogs.
\newblock {\em arXiv preprint arXiv:1510.03753}.

\bibitem[\protect\citename{Le and Mikolov}2014]{paragraph_vector}
Quoc Le and Tomas Mikolov.
\newblock 2014.
\newblock Distributed representations of sentences and documents.
\newblock In Tony Jebara and Eric~P. Xing, editors, {\em Proceedings of the
  31st International Conference on Machine Learning (ICML-14)}, pages
  1188--1196. JMLR Workshop and Conference Proceedings.

\bibitem[\protect\citename{LeCun \bgroup et al.\egroup }2015]{deep2}
Yann LeCun, Yoshua Bengio, and Geoffrey Hinton.
\newblock 2015.
\newblock Deep learning.
\newblock {\em Nature}, 521(7553):436--444.

\bibitem[\protect\citename{Lee \bgroup et al.\egroup }2015]{deepsupervision}
Chen-Yu Lee, Saining Xie, Patrick~W. Gallagher, Zhengyou Zhang, and Zhuowen Tu.
\newblock 2015.
\newblock Deeply-supervised nets.
\newblock In {\em AISTATS}, volume~38 of {\em JMLR Proceedings}. JMLR.org.

\bibitem[\protect\citename{Li \bgroup et al.\egroup }2015]{diversity}
Jiwei Li, Michel Galley, Chris Brockett, Jianfeng Gao, and Bill Dolan.
\newblock 2015.
\newblock A diversity-promoting objective function for neural conversation
  models.
\newblock {\em arXiv preprint arXiv:1510.03055}.

\bibitem[\protect\citename{Liu}2009]{ranking}
Tie-Yan Liu.
\newblock 2009.
\newblock Learning to rank for information retrieval.
\newblock {\em Foundations and Trends in Information Retrieval}, 3(3):225--331.

\bibitem[\protect\citename{Lowe \bgroup et al.\egroup }2015]{lowe2015ubuntu}
Ryan Lowe, Nissan Pow, Iulian Serban, and Joelle Pineau.
\newblock 2015.
\newblock The ubuntu dialogue corpus: A large dataset for research in
  unstructured multi-turn dialogue systems.
\newblock {\em arXiv preprint arXiv:1506.08909}.

\bibitem[\protect\citename{McClelland and Rogers}2003]{mcclelland}
James~L McClelland and Timothy~T Rogers.
\newblock 2003.
\newblock The parallel distributed processing approach to semantic cognition.
\newblock {\em Nature Reviews Neuroscience}, 4(4):310--322.

\bibitem[\protect\citename{Narasimhan \bgroup et al.\egroup }2015]{games}
Karthik Narasimhan, Tejas Kulkarni, and Regina Barzilay.
\newblock 2015.
\newblock Language understanding for text-based games using deep reinforcement
  learning.
\newblock In {\em Proceedings of the 2015 Conference on Empirical Methods in
  Natural Language Processing}, pages 1--11, Lisbon, Portugal, September.
  Association for Computational Linguistics.

\bibitem[\protect\citename{Parkinson \bgroup et al.\egroup }1977]{parry}
Roger~C Parkinson, Kenneth~Mark Colby, and William~S Faught.
\newblock 1977.
\newblock Conversational language comprehension using integrated
  pattern-matching and parsing.
\newblock {\em Artificial Intelligence}, 9(2):111--134.

\bibitem[\protect\citename{Ritter \bgroup et al.\egroup }2010]{twitterConv}
Alan Ritter, Colin Cherry, and Bill Dolan.
\newblock 2010.
\newblock Unsupervised modeling of twitter conversations.
\newblock In {\em Human Language Technologies: The 2010 Annual Conference of
  the North American Chapter of the Association for Computational Linguistics},
  HLT '10, pages 172--180, Stroudsburg, PA, USA. Association for Computational
  Linguistics.

\bibitem[\protect\citename{Schmidhuber}2015]{deep1}
J{\"u}rgen Schmidhuber.
\newblock 2015.
\newblock Deep learning in neural networks: An overview.
\newblock {\em Neural Networks}, 61:85--117.

\bibitem[\protect\citename{Schrading \bgroup et al.\egroup }2015]{reddit_abuse}
Nicolas Schrading, Cecilia Ovesdotter~Alm, Ray Ptucha, and Christopher Homan.
\newblock 2015.
\newblock An analysis of domestic abuse discourse on reddit.
\newblock In {\em Proceedings of the 2015 Conference on Empirical Methods in
  Natural Language Processing}, pages 2577--2583, Lisbon, Portugal, September.
  Association for Computational Linguistics.

\bibitem[\protect\citename{Serban \bgroup et al.\egroup
  }2015a]{serban2015building}
Iulian~V Serban, Alessandro Sordoni, Yoshua Bengio, Aaron Courville, and Joelle
  Pineau.
\newblock 2015a.
\newblock Building end-to-end dialogue systems using generative hierarchical
  neural network models.
\newblock {\em arXiv preprint arXiv:1507.04808}.

\bibitem[\protect\citename{Serban \bgroup et al.\egroup
  }2015b]{serban2015survey}
Iulian~Vlad Serban, Ryan Lowe, Laurent Charlin, and Joelle Pineau.
\newblock 2015b.
\newblock A survey of available corpora for building data-driven dialogue
  systems.
\newblock {\em arXiv preprint arXiv:1512.05742}.

\bibitem[\protect\citename{Shang \bgroup et al.\egroup }2015]{shang2015neural}
Lifeng Shang, Zhengdong Lu, and Hang Li.
\newblock 2015.
\newblock Neural responding machine for short-text conversation.
\newblock {\em arXiv preprint arXiv:1503.02364}.

\bibitem[\protect\citename{Sordoni \bgroup et al.\egroup
  }2015]{sordoni2015neural}
Alessandro Sordoni, Michel Galley, Michael Auli, Chris Brockett, Yangfeng Ji,
  Margaret Mitchell, Jian-Yun Nie, Jianfeng Gao, and Bill Dolan.
\newblock 2015.
\newblock A neural network approach to context-sensitive generation of
  conversational responses.
\newblock {\em arXiv preprint arXiv:1506.06714}.

\bibitem[\protect\citename{Sutskever \bgroup et al.\egroup }2014]{seq2seq}
Ilya Sutskever, Oriol Vinyals, and Quoc~V Le.
\newblock 2014.
\newblock Sequence to sequence learning with neural networks.
\newblock In {\em Advances in neural information processing systems}, pages
  3104--3112.

\bibitem[\protect\citename{Turing}1950]{turing}
Alan~M Turing.
\newblock 1950.
\newblock Computing machinery and intelligence.
\newblock {\em Mind}, 59(236):433--460.

\bibitem[\protect\citename{Vinyals and Le}2015]{techstop}
Oriol Vinyals and Quoc Le.
\newblock 2015.
\newblock A neural conversational model.
\newblock {\em arXiv preprint arXiv:1506.05869}.

\bibitem[\protect\citename{Walker \bgroup et al.\egroup }2012]{movie_chars}
Marilyn Walker, Grace Lin, and Jennifer Sawyer.
\newblock 2012.
\newblock An annotated corpus of film dialogue for learning and characterizing
  character style.
\newblock In Nicoletta Calzolari, Khalid Choukri, Thierry Declerck,
  Mehmet~U\u{g}ur Do\u{g}an, Bente Maegaard, Joseph Mariani, Jan Odijk, and
  Stelios Piperidis, editors, {\em Proceedings of the Eighth International
  Conference on Language Resources and Evaluation (LREC-2012)}, pages
  1373--1378, Istanbul, Turkey, May. European Language Resources Association
  (ELRA).
\newblock ACL Anthology Identifier: L12-1657.

\bibitem[\protect\citename{Wallace}2009]{alice}
Richard~S Wallace.
\newblock 2009.
\newblock {\em The anatomy of ALICE}.
\newblock Springer.

\bibitem[\protect\citename{Weizenbaum}1966]{eliza}
Joseph Weizenbaum.
\newblock 1966.
\newblock Eliza—a computer program for the study of natural language
  communication between man and machine.
\newblock {\em Communications of the ACM}, 9(1):36--45.

\bibitem[\protect\citename{{Wen} \bgroup et al.\egroup }2016]{deeptaskdialog}
T.-H. {Wen}, D.~{Vandyke}, N.~{Mrksic}, M.~{Gasic}, L.~M. {Rojas-Barahona},
  P.-H. {Su}, S.~{Ultes}, and S.~{Young}.
\newblock 2016.
\newblock {A Network-based End-to-End Trainable Task-oriented Dialogue System}.
\newblock {\em ArXiv e-prints}, April.

\bibitem[\protect\citename{Yao \bgroup et al.\egroup }2015]{yao2015attention}
Kaisheng Yao, Geoffrey Zweig, and Baolin Peng.
\newblock 2015.
\newblock Attention with intention for a neural network conversation model.
\newblock {\em arXiv preprint arXiv:1510.08565}.

\end{thebibliography}
\bibliographystyle{emnlp2016.bst}

\end{document}